# Simulating User Learning in Authoritative Technology Adoption: An Agent Based Model for Council-led Smart Meter Deployment Planning in the UK

Tao Zhang [a], Peer-Olaf Siebers [b], Uwe Aickelin [c]

[a] Department of Marketing,
Birmingham Business School, University of Birmingham

[b, c] Intelligent Modelling & Analysis Group,
School of Computer Science, University of Nottingham

Corresponding Author's Email: t.zhang.1@bham.ac.uk

## Abstract

How do technology users effectively transit from having zero knowledge about a technology to making the best use of it after an authoritative technology adoption? This post-adoption user learning has received little research attention in technology management literature. In this paper we investigate user learning in authoritative technology adoption by developing an agent-based model using the case of council-led smart meter deployment in the UK City of Leeds. Energy consumers gain experience of using smart meters based on the learning curve in behavioural learning. With the agent-based model we carry out experiments to validate the model and test different energy interventions that local authorities can use to facilitate energy consumers' learning and maintain their continuous use of the technology. Our results show that the easier energy consumers become experienced, the more energy-efficient they are and the more energy saving they can achieve; encouraging energy consumers' contacts via various informational means can facilitate their learning; and developing and maintaining their positive attitude toward smart metering can enable them to use the technology continuously. Contributions and energy policy/intervention implications are discussed in this paper.

## 1. Introduction

Technology adoption (or Innovation diffusion) theories focus on understanding how, why and at what rate innovative ideas and technologies spread in a social system (Rogers, 1962). In technology adoption processes the decisions of whether to adopt an innovative technology can either be made by the actual users freely and implemented voluntarily, or be made by a few authoritative individuals and implemented enforcedly.

In the former type of technology adoption, it is usually assumed that before an actual user makes the adoption decision of a particular technology, he/she has learned some knowledge or even gained some experience about it (e.g., the information search stage in the five-step consumer decision model (Engel et al, 1995)).

In the latter type of technology adoption, once the adoption decision has been made the actual users would be "forced" to use a technology with very limited prior knowledge/experience about it. This type of technology adoption usually takes place at the level of a massive system or infrastructure upgrade. An example for such a case is a university-wide systematic upgrade of the office and lab computer operating system from Windows XP to Windows 7. In this case, the decision is made by the management of the university, and the actual users (e.g. faculty staff and students) are forced to use the innovation with limited or even no knowledge about it beforehand and no influence on the choice[1].

Whilst free adoption decisions and voluntary use in innovation diffusion received intensive studies (e.g., Griliches, 1957; Mansfield, 1961; Rosenberg, 1972; Geroski, 2000; Hall & Khan, 2002), authoritative adoption decisions and forced use in innovation diffusion seems to be an area in its infancy stage. An innovation cannot benefit the society unless its actual users use it effectively. Thus when an authoritative adoption happens, it is significantly important to understand how actual users start to learn about the innovative technology, use the technology, and finally make the best use of it and perhaps motivate other users to use it or to improve their knowledge quickly. Users' transition from having zero knowledge about a technology to making the best use of it is a consumer learning process. In technology deployment planning, understanding this learning process would help decision-makers design strategies to accelerate users' transition, maintain users' interest in the technology and maximise the benefits that the technology can bring to the society.

---

[1] We acknowledge that in technology adoption (or innovation diffusion) studies there is a concept "induced diffusion", which has been defined as "any intervention that aims to alter the speed and/or total level of adoption of an innovation by directly or indirectly internalising positive and/or negative externalities" (Davies & Diaz-Rainey, 2011; p.1229). Induced diffusion research primarily investigates how the diffusion of new technologies can be altered by policy interventions, e.g. economic incentives, information provision or regulations (Diaz-Rainey, 2009). The preponderance of induced diffusion studies use economic modelling approaches based on firm-level data to examine the macro-level patterns of technology diffusion (Diaz-Rainey, 2009). These studies do not look at the adoption decision-making and post-adoption learning behaviour of individual adopters. As noted by Diaz-Rainey (2009, p.20), "there is clearly a need to understand whether inducing diffusion among individuals is substantially different to doing so among multinational corporations". The term "authoritative adoption" in our paper is defined as "a technology adoption where the adoption decision is made by a few authoritative individuals and implemented enforcedly, and the actually users are forced to use the technology", which is different from "induced diffusion". Thus the positioning of the paper is "post-adoption user learning in authoritative technology adoption".

Traditionally, there are some theories aiming to understanding how users (or consumers) learn and adopt a technology, for example, the Technology Adoption Model (TAM) (Davis, 1989) in information systems research and consumer learning models (Solomon et al, 1999) in consumer research. Many of these theories/models are qualitative, static, and only apply to adoption decisions that are made on voluntary basis. In other words, users/consumers seek information and learn knowledge about the product/innovation on their own, and then make purchase decisions voluntarily.

Currently no studies extend their application to authoritative technology adoptions. In this paper, we bridge this academic gap: we draw on the ideas from consumer learning theories/models, and extend the application of them to authoritative technology adoptions by developing a computational model using Agent-Based Simulation (ABS).

The agent-based model we have developed is based on a case of smart meter deployment in the UK City of Leeds. This case provides a good example of an authoritative technology adoption: the city council uses smart metering energy intervention to systematically upgrade the energy infrastructure in the city, and some energy users (i.e. those who live in council-owned properties) will have smart meters installed at their homes and are forced to use them. With the simulation model we would like to visualise the dynamic process of user learning and understand effects of the learning process on making the best use of the technology (i.e., when users make the best use of smart metering technology, they are on effective electricity demand side management, which we can use electricity consumption data to monitor). Similar studies in *Technological Forecasting & Social Change* (e.g. Gordon 2003; Schwarz & Ernst, 2009; Rixen & Weigand, 2014) have proved that agent-based simulation is an effective approach for studying various areas of technology adoption.

This paper serves two purposes. Academically we want to advance the academic knowledge in technology management by studying for the first time the field of authoritative technology adoption by extending the application of consumer learning theories in conjunction with empirical data to that field via a computational simulation method. Practically we aim to develop a smart meter development planning tool (i.e. a software decision support platform) to provide hands on advice to city council decision-makers with policy implications on how effectively to facilitate user learning and maximise the benefits of smart meters for the city.

The paper is structured as follows. In the second section we review relevant theories about consumer learning. In the third section we describe the case study and the simulation model and its individual components in detail. In the fourth section we carry out four simulation experiments, present the experiment results and their related implications. The fifth section discusses the study, and the sixth section concludes the study.

## 2. Theoretical Background

Amongst the researchers studying learning processes there is no consensus about how learning happens. Thus the definition of learning is diverse. In psychology, researchers view learning as a relatively permanent change in behaviour as a result of increasing experience (Solomon et al, 1999). In marketing, consumer learning is defined as "a process by which individuals acquire the purchase and consumption knowledge and experience they apply to future related behaviour (Schiffman et al, 2008, p. 185). In the real world, individuals learn both directly and indirectly. For example, they can learn from the events that directly influence them; or they can learn from other people's experiences indirectly; sometimes they even learn unconsciously.

Learning covers activities ranging from consumers' responses to external stimuli to a complex set of cognitive processes. There are many learning theories which generally fall into two categories: behavioural learning and cognitive learning.

### 2.1 Behavioural Learning

The behavioural learning approach makes the assumption that learning happens as a result of responses to external stimuli (Solomon et al, 1999). Thus sometimes behavioural learning theories are also known as stimulus-organism-response (SOR) theories, as these theories primarily focus on the inputs and outputs in the learning process. The behavioural approach takes the view that a learner's mind is a "black" box, and emphasizes the observable perspectives of behaviour, as shown in Figure 1.

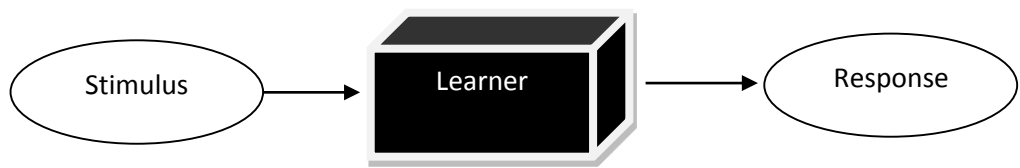

Figure 1: The learner as a "black box" in behavioural learning (Solomon et al, 1999)

In behavioural learning theories learners are mindless passive objects, i.e. they do not make decisions; they can only be taught certain behaviour through repetition or conditioning (Schiffman et al, 2008).

A quantitative expression of the behavioural approach was developed in early marketing literature, e.g. Estes (1950, cited in Bennett and Mandell (1969)), Estes and Buke (1953, cited Bennett and Mandell (1969)), and Bush and Mosteller (1955, cited in Bennett and Mandell (1969)). In all cases learning is treated as a stochastic process and thus response tendencies are treated in probabilistic terms. Howard (1963, cited in Bennett and Mandell (1969)) proposes the following consumer's brand choice learning function:

$$P_A = M(1 - e^{-kt}) \quad (1)$$

Where $P_A$ is the probability of response (i.e. purchasing Brand A); $M$ is the maximum attainable loyalty to Brand A; $k$ is the constant that expresses the learning rate; $t$ is the number of reinforced trials.

This quantitative model (see Figure 2) was empirically validated in Bennett and Mandell (1969) by using the case of new automobile purchase.

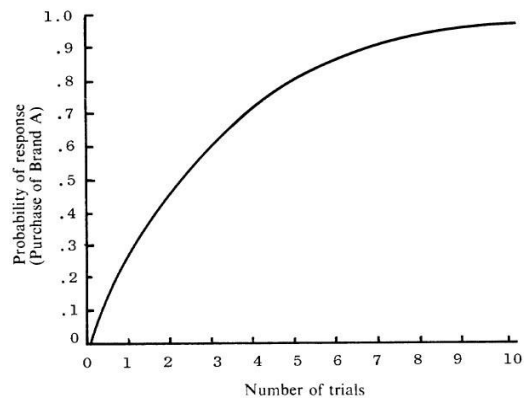

Figure 2: The learning curve for Brand A (Source: Bennett and Mandell, 1969)

### 2.2 Cognitive Learning

The cognitive learning approach assumes that learning is a set of mental processes. In contrast to the behavioural learning approach described earlier the cognitive learning approach takes the view that learners are problem-solvers rather than "black boxes". In other words, learners make purchase/adoption decisions on their own rather than passively repeat trial behaviour. They actively seek information about a product/innovation, process the information and gain motivation or intention to buy/adopt the product/innovation. A typical cognitive learning theory is observational learning theory, which believes that "individuals observe the actions of others and note the reinforcement they receive for their behaviours" (Solomon et al, 1999, p.70). The entire process of observational learning is presented in Figure 3.

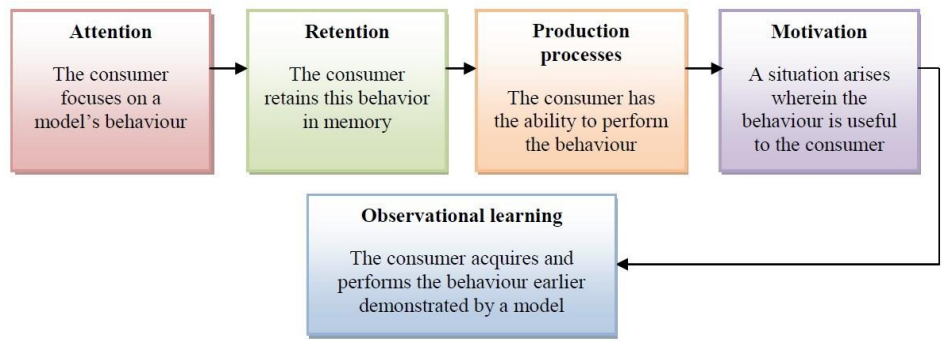

Figure 3: Process of observational learning (Source: Solomon et al, 1999)

**2.3 Our Choice**

Through a review of the two learning approaches we can see that, whilst both approaches are useful and have been empirically validated, a main distinction between the two is whether users/consumers are passive recipients or active decision-makers. Since the study in this paper focuses on forced authoritative technology adoption in which users passively receive the technology and are taught to use, we draw on the ideas of behavioural learning approach to develop the simulation model.

**3. The Model**

**3.1 The Case: Smart Metering Energy Interventions in Leeds**

As climate change has become a very important global issue, the UK government has set a national target: cutting $CO_2$ emission by 34% of 1990 levels by 2020. In the UK there are nearly 70 cities and it is increasingly being suggested that each city will have an important role to play for the country's future sustainability (Keirstead and Schulz, 2010; Bale et al., 2012; Zhang et al., 2012). With a population of 787,700 (Office for National Statistics, 2011), Leeds City Council (Leeds CC) is the second largest metropolitan council in England and also the UK's largest centre for business, legal and financial services outside London.

Having realised its important role for the future UK sustainability, Leeds CC made a clear statement to voluntarily undertake a target of cutting its $CO_2$ emission by 40% by 2020. However, like many other UK city councils, Leeds CC has no explicit local energy policies/interventions for achieving this target so far. Currently Leeds CC is working with the Universities of Leeds and Nottingham and the research council (EPSRC) on a *City Energy Future* project that aims to seek knowledge, experience and develop decision support tools to aid local energy intervention design. Several local energy interventions have been proposed, including setting up a city level strategic council, developing local district heating networks, running energy saving education campaigns and deploying smart meters.

Smart metering is a popular area in energy policy research. There is a nuanced debate over the benefits/drawbacks of this technology. On the positive side, smart metering has been empirically proved to be an effective means to influence energy users' behaviour and reduce energy consumption (e.g. Haney et al., 2009; Martiskainen & Ellis, 2009; DECC, 2012; Xu et al., 2015). On the negative side there are critiques that smart meters are just cost-saving devices for utility companies and can actually make energy consumers use more energy; there are many security issues with smart metering (Sharma and Saini, 2015). In the UK there are also campaigns against the rollout of smart meters (e.g. Stopsmartmeters!). Clearly smart metering as an innovative technology cannot benefit the society without energy consumers' behaviour change (Owens & Drifill, 2008). How much behavioural change can smart meters lead to is a concern when deploying smart meters has been considered to be an energy intervention (Martiskainen & Ellis, 2009). Smart metering intervention has been proposed for Leeds CC because in the UK context there is empirical evidence that smart metering can lead to effective energy consumers' behaviour change and substantial energy saving (Haney et al., 2009; Martiskainen & Ellis, 2009). Additionally, smart metering can also help energy consumers in the UK alleviate fuel poverty[2] (Barnicoat & Danson, 2015). Many residents living in council-owned properties in Leeds are considered to be in fuel poverty. As Leeds CC has direct control over the council-owned properties, authoritatively installing smart meters in the council-owned properties would potentially be an effective way to help the occupants get out of fuel poverty (through changing their behaviour of using energy thus reducing their energy consumption and cutting their energy bills).

**3.2 Modelling Rationale**

The research in the paper targets the case of authoritative smart metering adoption in Leeds by developing a computational model using Agent-Based Modelling and Simulation (ABMS). ABS is a computational modelling approach to study Multi-Agent Systems (MAS). An Agent-Based Model (ABM) is composed of individual agents, commonly implemented in software as objects. Agent objects have states and rules of behaviour. Furthermore they often have a memory and the capability to evolve over time. This modelling approach lends itself particularly well for modelling people and their behaviour (Siebers and Aickelin, 2011). Running an agent-based model simply amounts to instantiating an agent population, letting the agents behave and interact, and observing what happens globally (Axtell, 2000). Thus a unique advantage of ABS is that almost all behavioural attributes of agents can be captured and modelled. ABS is widely adopted in studying MAS, particularly those with intelligent human beings (e.g. markets, societies, and organisations). In this particular case, the energy users behave and interact in a community, which is a complex social system that is well suited to agent-based simulation.

---

[2] A household is said to be in fuel poverty if it needs to spend more than 10% of its income on fuel to maintain a satisfactory heating regime (usually 21 degrees Celsius for the main living area, and 18 degrees Celsius for other occupied rooms) (UK National Statistics, 2011)

### 3.3 Model Design

We draw on the idea of the consumat[3] approach (Janssen and Jager 1999; Jager 2000) and model residential energy consumers (REC) as intelligent agents. Instead of simply adopting the consumat approach, we modify it by considering the empirical evidence from the electricity market. A key difference between the original consumat approach and our modified version is that: the original consumat is developed based on a combination of some well-established psychological theories such as human needs, motivational processes, social comparison theory, theory of reason action, etc., while our modified version is developed based on an empirical survey of the residential electricity consumers in the real electricity market.

A template of our REC agents and an overview of the model are shown in Figure 4. According to our previous research (Zhang et al., 2011, 2012), we consider a household's base electricity consumption which is determined by the household basic energy need, and flexible electricity consumption which is determined by the greenness of the occupants' behaviour and the number and types of electrical appliances in the house. The household basic energy need is influenced by the number of occupants, type of property and tenure. The greenness of the occupants' behaviour is determined by their energy-saving awareness and their attitude toward using energy-saving technologies. The greener the occupants' behaviour is, the higher the probabilities that they respond to the information provided by smart meters (i.e. switching off the unnecessary appliances when the electricity price is high). The REC agents also interact with each other. Their interactions can change their energy-saving awareness and attitude. Property occupying times determines whether the occupants' green behaviour is effective or not. For example, if the occupants are at home, they can switch on/off the appliances; when they are out of home they cannot do this.

---

[3] The "consumat approach" is a model of human behaviour with a particular focus on consumer behaviour. It is based on concepts and theories from psychology, economics and computer science. The conceptual framework is a kind of meta-model of the many theories in psychology. The computational version of the consumat approach is based on multi-agent simulation. For details please see Janssen and Jager (1999) and Jager (2000).

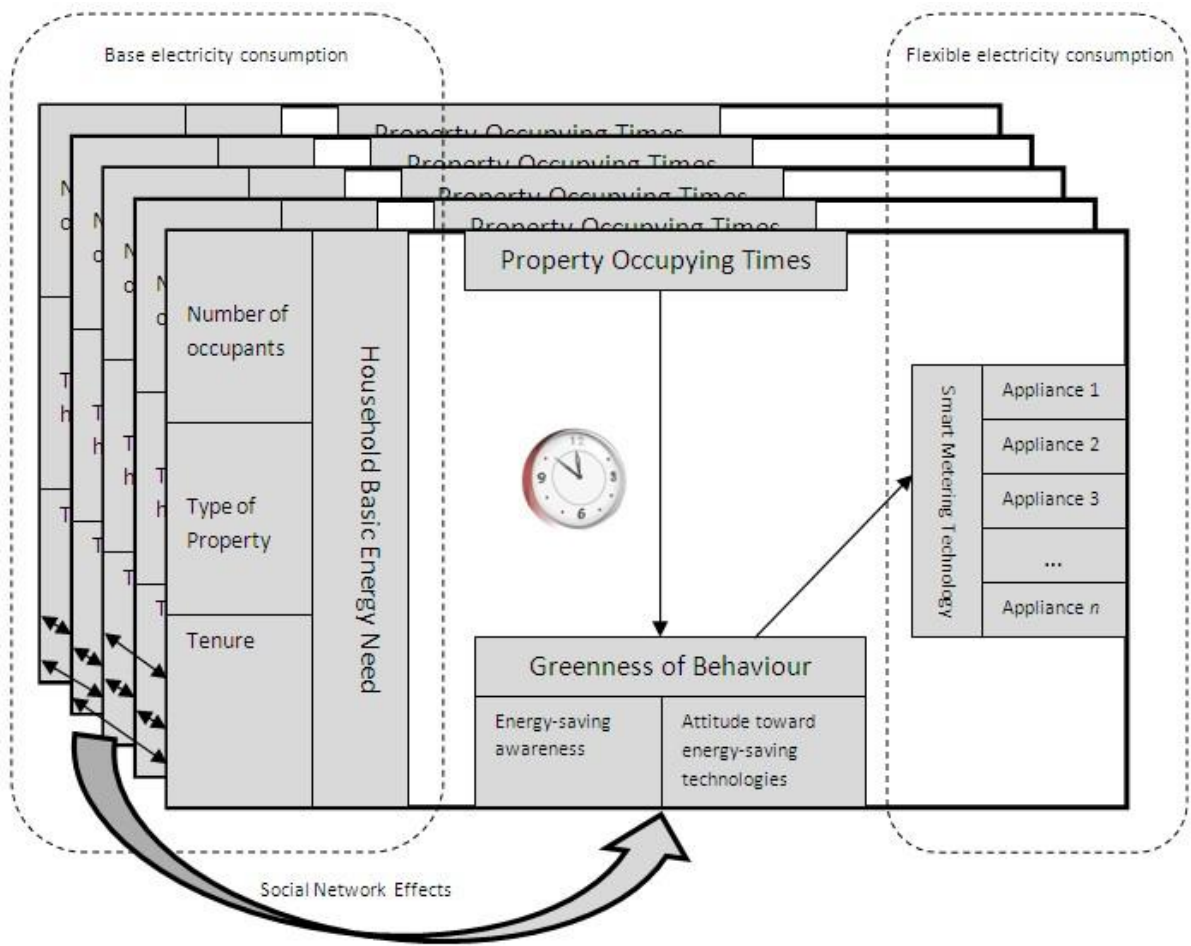

Figure 4: The overall view of the model

### 3.3.1 Behaviour of Residential Energy Consumer Agents

We use a state chart to represent the behaviour of REC agents, as shown in Figure 5. Each REC agent has a set of home electrical appliances. We have carried out a survey to get empirical data about people's social demographic attributes, number and types of electrical appliances, levels of awareness about energy, and life styles in the Leeds area. Based on the empirical data, we have developed some energy consumer archetypes (Zhang et al. 2012) shown in Table 1. These REC agent archetypes determine the initial parameter settings in the simulation.

Table 1: UK residential energy consumer archetypes (Zhang et al. 2012)

| Archetype | Attributes | | |
|---|---|---|---|
| | Property energy efficiency level | Greenness of behaviour | Duration of daytime occupancy |
| 1: Pioneer Greens | High | High | Short |
| 2: Follower Greens | Low | High | Short |
| 3: Concerned Greens | Low | High | Long |
| 4: Home-Stayers | High | High | Long |
| 5: Unconscientious Wasters | High | Low | Short |
| 6: Regular Wasters | Low | Low | Short |
| 7: Daytime Wasters | High | Low | Long |
| 8: Disengaged Wasters | Low | Low | Long |

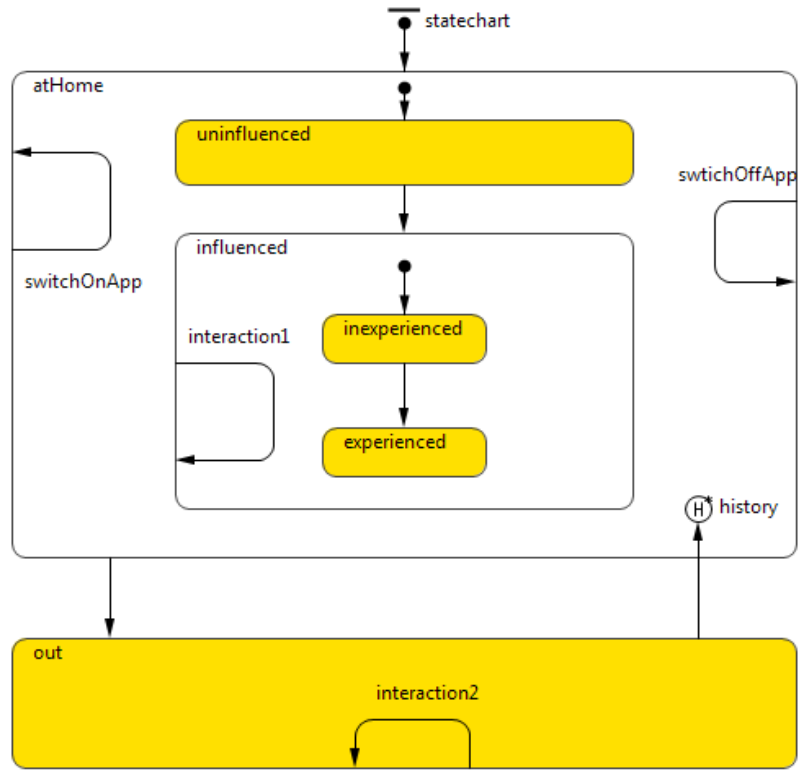

Figure 5: The state chart of REC agents

In the model we simulate the daily life of energy consumers. An REC agent has two main states: "atHome" and "out". The transition from "atHome" state to "out" state is controlled by the parameter "timeLeaveHome", and the transition from "out" state to "atHome" state is controlled by the parameter "timeBackHome". Both parameters are defined by the archetype of the REC agent. For example, a REC agent of "short daytime occupancy" (e.g., a regular work professional) often leaves home for work at a time between 8:30am to 9:30am, and goes back home at a time between 5:30pm to 6:30pm. These two transitions reflect the "Property Occupying Times" in Figure 4.

When the REC agent is in the "atHome" state, it initially is in an "uninfluenced" sub-state if the city council does not implement the smart metering intervention. Otherwise the REC agent transits from the "uninfluenced" sub-state to the "influence" sub-state. When an REC agent is influenced by the city council's smart metering intervention it initially is in the "inexperienced" sub-sub-state. Then the REC agent starts the learning process by trying to use the smart meter. Drawing on the idea of the learning theory chosen in Section 2.3, in the simulation model we re-interpret the learning function. We interpret $P_t$ as an REC agent's probability of responding to the information provided by the installed smart meter; $M$ as the REC agent's attitude ($A$) toward smart metering technology; $K$ as the REC agent's energy-saving awareness ($ESA$); and $t$ is the number of tries. Thus the REC agent's learning function can be defined as:

$$P_t = A * (1 - e^{-(ESA)*t}) \quad (2)$$

From eq. 2 we can see that the more an REC tries the technology, the higher the probability that it responds to the information provided by a smart meter. We posit that an REC agent's experience in using smart metering technology is reflected by the probability that it responds to the information provided by a smart meter. The transition from the "inexperienced" sub-sub-state to the "experienced" sub-sub-state is a probability threshold ($P_{th}$), which in the simulation model is initially arbitrarily set at 0.85. The probability ($P_t$) is calculated according to the learning function (eq. 2). If $P_t$ is larger or equal to $P_{th}$, the REC agent makes the transition from the "inexperienced" sub-sub-state to the "experienced" sub-sub-state. The value of $P_{th}$ reflects the level of experience energy consumers need to gain when they make the best use of smart meters in the real world. The smaller the value of $P_{th}$ is, the easier the energy consumers transit from the "inexperienced" sub-sub-state to the "experienced" sub-sub-state, thus achieving energy-efficiency. We also set the probability threshold ($P_{th}$) at 0.8 and 0.9 levels to check the sensitivity of the simulation results.

When an REC agent is at home, it can switch on/off appliances. This behaviour is reflected by the two internal transitions "switchOffApp" and "switchOnApp". The initial probabilities for them to occur are determined by its archetype. Once the REC agent has a smart meter, the probabilities for them to occur are determined by $P_t$ calculated by the learning function, i.e., if the REC agent responds to the information provided by the smart meter, it changes its energy use behaviour accordingly.

When an REC agent is influenced by the smart metering energy intervention, it can interact with other REC agents through a network to exchange the knowledge and experience. This is reflected in the state chart by the two internal transitions "interaction1" and interaction2". These two internal transitions can influence the REC agent's attitude ($A$) and energy saving awareness ($ESA$) positively or negatively, as shown by the social network effects in Figure 4. This reflects a reality that in the real energy market sharing knowledge and experience via communications can change consumers' attitude and energy saving awareness (Owens and Driffill, 2008). In the model, we have chosen a small world social network to model the communication channels between REC agents.

### 3.3.2 Model Implementation

We use AnyLogic 6.7.1 (XJ Technology, 2012) to develop the agent-based model. We use the real map of the Leeds Metropolitan District Area as a background to design the model, as shown in Figure 6. The model has been implemented on a standard PC with Windows XP SP3.

We develop 1000 REC agents to simulate 1000 residential energy consumers living in council-owned properties. According to our survey, residential energy consumers living in council-owned properties are of archetypes "follower greens", "concerned greens", "regular wasters" and "disengaged wasters". Based on the survey statistics, we set "follower greens" 11%, "concerned greens" 13%, "regular wasters" 47%, and "disengaged wasters" 29%.

We set each time step in the simulation model as one minute, and simulate the daily life of residential energy consumers and observe and analyse how their learning can result in the dynamics of electricity consumption at the community level. An overview of the model is shown in Figure 6, where the blue dots are REC agents randomly distributed on the map. When the simulation begins, the REC agents are initial blue. When we choose "smart metering", they turn to red, which means they have been affected by the city council's smart metering energy policy, i.e. the city council has installed smart meters into these council-owned properties. Then the REC agents start the learning process defined by the learning function (eq.2). They also start to interact and the interactions will influence their learning. With their learning over time, some REC agents turn from red to yellow, which means they transit from "inexperienced" to "experienced" in using smart meters. The parameter settings in the simulation are listed in Table 2. In order to enable other researchers to use the model, we uploaded the model online at http://www.runthemodel.com/models/879/

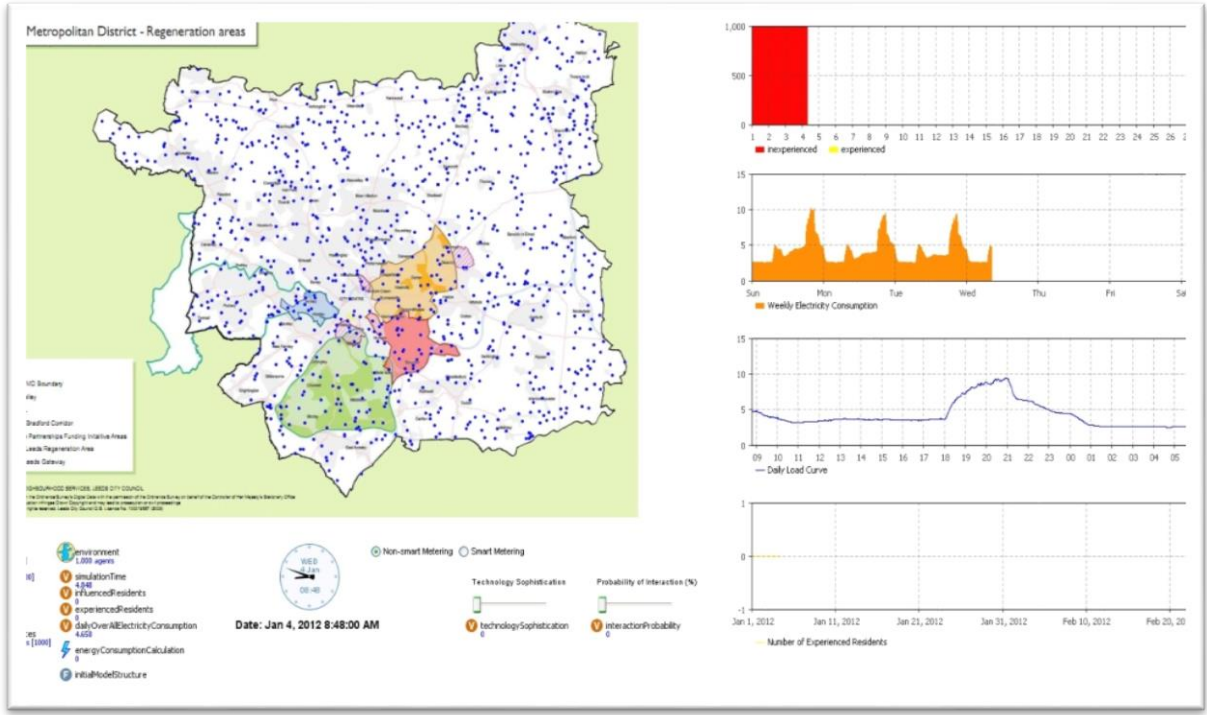

Figure 6: Interface of the model

Table 2: Simulation parameter settings

| Agent Archetype | Percentage | A | ESA | timeLeaveHome | timeBackHome | $P_{th}$ |
|---|---|---|---|---|---|---|
| Follower greens | 11% | Normal distribution (μ=0.71, σ=0.052) | Normal distribution (μ=0.74, σ=0.041) | 6:00 to 9:00, random uniform distribution | 15:00 to 18:00, random uniform distribution | 0.85 |
| Concerned greens | 13% | Normal distribution (μ=0.69, σ=0.050) | Normal distribution (μ=0.72, σ=0.043) | 9:00 to 18:00, random uniform distribution | timeLeaveHome + random (1, 180 min) | 0.85 |
| Regular wasters | 47% | Normal distribution (μ=0.39, σ=0.061) | Normal distribution (μ=0.41, σ=0.033) | 6:00 to 9:00, random uniform distribution | 6:00 to 9:00, random uniform distribution | 0.85 |
| Disengaged wasters | 29% | Normal distribution (μ=0.22, σ=0.037) | Normal distribution (μ=0.25, σ=0.057) | 9:00 to 18:00, random uniform distribution | timeLeaveHome + random (1, 180 min) | 0.85 |

## 4. Experimentation

In order to study the effects of user learning in authoritative technology adoption, we carried out four experiments. Drawing on Rixen and Weigand (2014), we use key performance indicators (KPIs) to show the experiment results. As the purposes of these experiments are different, they have different KPIs.

Experiment 1: Validating the model

The first experiment is meant for model validation. The KPI we use in the experiment is the average daily electricity load curve (kW). We run the model 50 times and gained the average daily load curve (half hourly) of the whole virtual community (1000 REC agents in total) during winter time. We then calculate the average daily load curve of an individual REC agent and compare the result with the real standard domestic load curve provided by the Electricity Association (Abu-Sharkh et al., 2006). The comparison in Figure 7 shows that the two patterns of domestic load curves are quite similar to each other and improves our confidence in the validity of the simulation model.

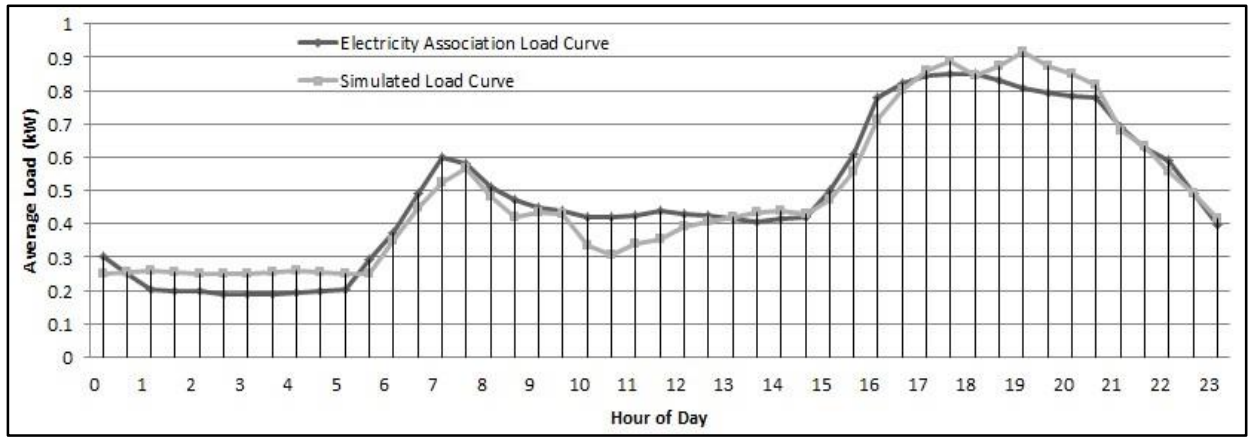

Figure 7: Simulated load curve vs. Real load curve

Experiment 2: Agent transition from "inexperienced" to "experienced"

In the second experiment, we simulated the REC agents' transition from the "inexperienced" to the "experienced" sub-sub-state. The purpose of the experiment is to show the effect of user learning on the community's energy consumption. In this experiment we again use the average daily electricity load curve (kW) as the KPI. We run the model, and gain domestic average load curves in both the "inexperienced" scenario and the "experienced" scenario. In the "experienced" scenario, we set the probability threshold ($P_{th}$) at 0.8, 0.85 and 0.9 levels to check the sensitivity of the simulation results, as shown in Figure 8. In the "inexperienced" scenario all the REC agents are inexperienced, while in the "experienced" scenario 80% of the REC agents have transited from "inexperienced" to "experienced" in using smart meters. Each curve is the average result of 50 runs of the simulation model. The comparison in Figure 8 shows that the REC agent's transition from "inexperienced" to "experienced" can cause substantial reduction of energy consumption at peak times. This simulation finding has been evidenced by empirical observations in the real electricity market (Haney et al., 2009). The reductions of energy consumption at different $P_{th}$ levels are different. When the $P_{th}$ is at 0.8, 0.85 and 0.9 levels, the average daily energy reduction is 19.5%, 18.9% and 18.3% respectively. This simulation finding reflects that in the real world, if energy consumers can more easily become experienced in using smart meters they tend to become energy-efficient quicker and achieve more energy saving. The simulation finding also means that facilitating user learning can enable users to make better use of an innovative technology thus gaining more benefit from the technology.

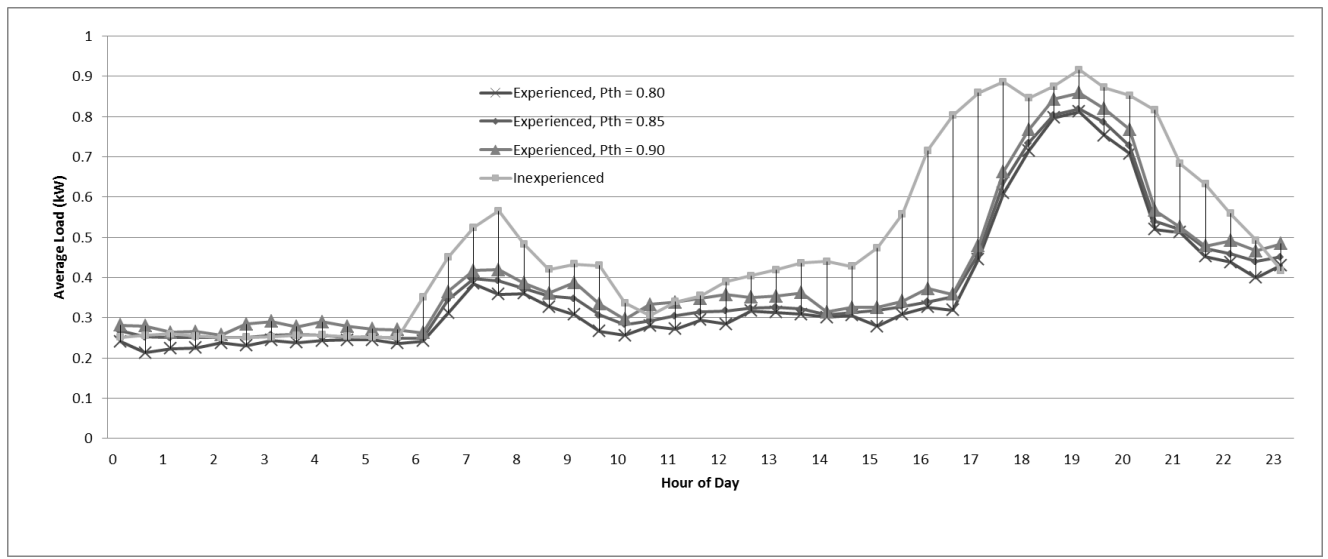

Figure 8: "inexperienced" vs. "experienced"

Experiment 3: Learning facilitation

In the third experiment, we study the social network effects on user learning. The purpose of the experiment is to explore the strategies that can facilitate user learning. We use the number of experienced users as the KPI. We adopt a small-world (a built-in social network in AnyLogic 6.7.1) and set a contact rate for the REC agents. The contact rate is a probability with which an REC agent contact other REC agents who are connected with it in the social network. When an REC agent contacts other REC agents, it sends a message to them. The message can change other REC agents' energy-saving awareness (***ESA***) and attitude (***A***). We set the contact rate 0.5, 0.3 and 0.1, and plot the transition trends over three months (90 days virtual time) in Figure 9 (each curve is the average result of 50 runs). From Figure 9 we can see that contact rate has significant influence on the transition. The higher the contact rate is, the faster the transition is and the more the final experienced users there are. The simulation finding means that in the real world when an authoritative technology adoption happens, encouraging users to share experience and exchange knowledge about the technology will facilitate user learning.

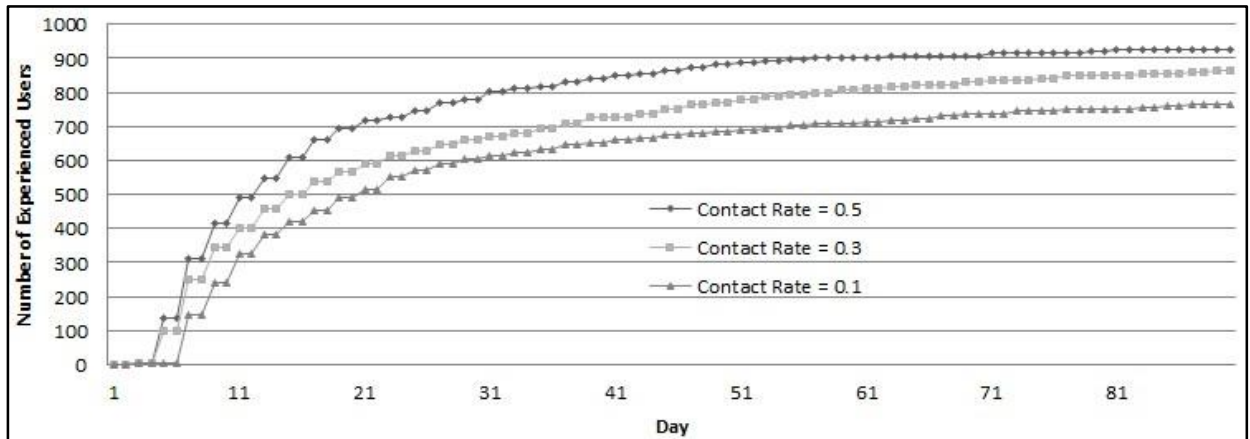

Figure 9: The influence of contact rate on transition

Experiment 4: Technology discontinuance

In the fourth experiment, we focus on the REC agents' continuation of using smart meters. If all the REC agents can quickly learn how to use smart meters effectively and maintain their interest in using the technology, they can maximising the benefits of the technology. Technology continuance is a very interesting topic in information systems and marketing studies. There are some empirical studies focusing on identifying the factors that can influence users' continuation of using innovative technologies (e.g. Spiller et al, 2007; Parthasarathy and Bhattacherjee, 1998; Karahanna et al. 1999). In this simulation study, according to the learning function (eq.2) an REC agent's probability of responding to the information provided by smart meters (*P*) is influenced by its attitude (*A*) and energy-saving awareness (*ESA*). Because of the social network effects, both *A* and *ESA* are changing over time. By exploring how the *A* and *ESA* of discontinuers evolve, we can find out how significant they are in technology discontinuance.

We stick to the transition probability threshold ($P_{th}$ = 0.85). We assume that after an REC agent transits from "inexperienced" to "experienced" sub-sub-state (*P* >= $P_{th}$), if its *P* later drops below $P_{th}$, it loses its interest in using its smart meter and becomes a discontinuer (however it will not transit back from "experienced" to "inexperienced" as in the real word a user can lose its interest in using a technology although he/she is still experienced in using it).

Using the *A* and *ESA* as the KPIs, we run the model 50 times (90 days virtual time) and each time we can identify a small proportion (8%) of discontinuers. We collect the statistics of both continuers and discontinuers' *A* and *ESA* and calculate the average of an individual, as shown in Figures 10 and 11. From a comparison between the two figures we can see that whilst both continuers and discontinuers have the same trend in their *ESA*, a key difference between them is their *A*.

Discontinuers drop their *A*s significantly. The simulation finding shows that attitude is one significant factor influencing users' continuation in using a technology. A major reason for users to discontinue to use a technology is that they change their attitude negatively (i.e. decreasing *A*s). This simulation finding is also in line with some empirical research in information systems studies (e.g. Karahanna et al. 1999).

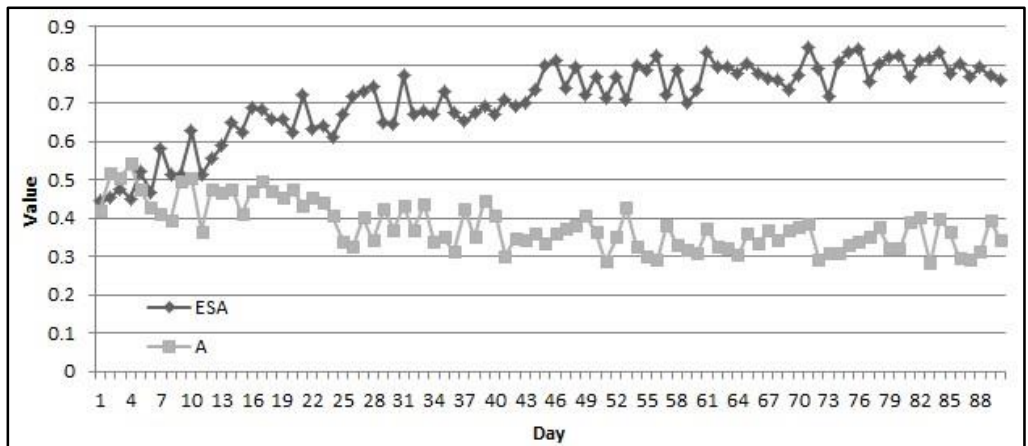

Figure 10: The average values of individual discontinuers' *A* and *ESA*

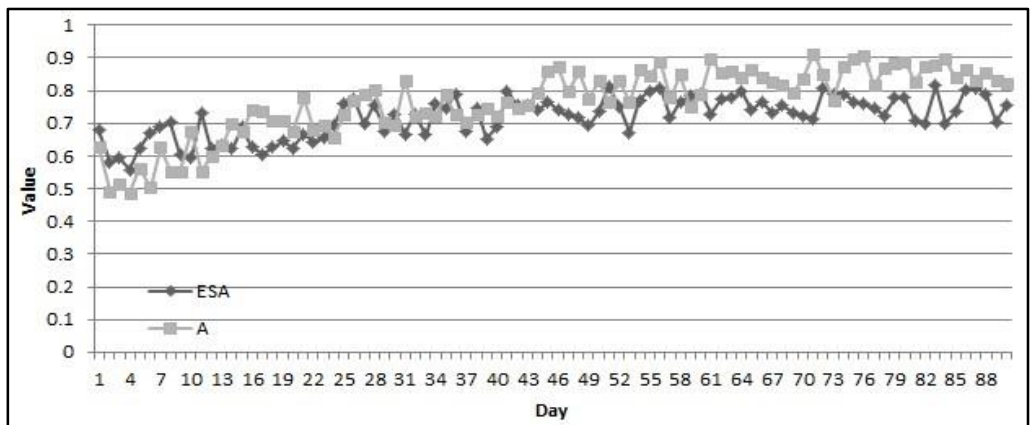

Figure 11: The average values of individual continuers' *A* and *ESA*

## 5. Discussions

**5.1 Advantages of the study**

This study—for the first time—develops an agent-based model to study the effects of user learning in authoritative technology adoption, an area that has been ignored in most other studies. Authoritative technology adoption is quite common, particularly when the authorities systematically upgrade utility infrastructures in communities. Thus studying authoritative technology adoption can help authority decision makers make better strategies for maximising the benefits of adopted technologies. In this particular study, we used the case of smart meter deployment in the UK City of Leeds. A unique advantage for using this particular case is that the benefit of the technology (i.e. smart metering) can be quantatively presented by the reduction of electricity consumption, which we can easily collect with the simulation model (e.g. the energy savings at different levels in Experiment 2).

Our choice of the agent-based simulation method enables us to capture the learning behaviour of users at micro level well. The REC agents were created based on the residential energy consumer archetypes developed from our empirical survey in the City of Leeds. We have also based the REC agents' learning behaviour on the behavioural learning theory and the learning function which were extensively justified by empirical studies (e.g. Bennett and Mandell, 1969; Schiffman et al, 2008; Solomon et al, 1999). Thus the REC agents have a solid empirical grounding. We also validated our simulation output against the real world observation (i.e. real domestic load profile from the Electricity Association). We believe the research outputs are valid and reliable and can provide meaningful decision implications.

**5.2 Energy policy/intervention implications**

In 2009 the UK Government announced a policy of installing smart meters for all residential energy consumers in GB by 2020. The latest statistical release shows that by 30 September 2014 a total number of 621,600 domestic smart meters were installed, and around 543,900 smart meters are now operating in smart mode in domestic properties in GB (DECC, 2014). It is expected that city councils play significant roles in the rollout process (especially the installations of smart meters in council-owned properties). From the study we can see that when authoritative installations of smart meters take place in council-owned properties, enabling energy users quickly to gain experience can be an effective means to reap the benefit of smart metering. Energy users' quicker and easier transitions from the "inexperienced" sub-sub-state to the "experienced" sub-sub-state (i.e. reflected by a smaller $P_{th}$ value) can result in substantial energy saving. This learning process can be facilitated by encouraging contacts via various informational means. For example, city councils can run energy educational programmes, events or energy saving campaigns etc. for council house residents, which can encourage their contacts for sharing knowledge and experience about smart metering and energy saving. Moreover, after the installations it is significantly important to develop and maintain energy users' positive attitude toward smart meters so as to encourage them to use the technology continuously. This can be achieved through, for example, using council leaflets to highlight the energy savings and environmental benefits they gain from smart metering.

**5.3 Limitations of the study**

Although the REC agents developed in the research have a strong empirical root, they cannot perfectly replicate the residential energy consumers in the real world. Thus it is important to acknowledge the limitations of the research. A first limitation is related to the critique of the behavioural learning theory which sees users as mindless passive objects. Some people believe that although actual users passively accept a technology in authoritative technology adoption, they still need to actively learn how to use it. The current simulation model in the research cannot capture the users' activeness. A second limitation is that in the simulation model the REC agents' archetypes are fixed. In other words, in the simulation there is no way for the REC agents to switch their archetypes. But we note that in the real world the change of archetypes can happen (for example, if a residential energy consumer loses his/her job, he/she might spend more time at home), although the probability for its occurrence is low. A third limitation is related to the critique of using simulation models as decision support tools to guide practice. Numbers and graphs from simulations may impress local authorities and the public, which can help to boost their receptivity to further dialogues and collaboration. However, this is just the beginning. The true utility of the method will only be proven in practice.

## 6. Conclusions and Future Research

In this paper we described an agent-based model for studying the effects of user learning on authoritative technology adoption. We began the paper with an argument for the importance of user learning in authoritative technology adoption, and then reviewed two important consumer learning theories and justified our choice of the behaviour learning theory in this particular study. We then developed an agent-based model to study user learning in authoritative technology adoption based on the case of smart meter deployment in the UK City of Leeds, and presented the simulation outputs and their related implications. Along the way, we focused on two objectives. One (in innovation management and marketing) is combining empirical data and a well-established theory to develop an agent-based model in response to a real world problem. The other one is presenting a first draft of a smart meter deployment planning tool and reusable residential energy consumer templates for studying issues in the energy market.

The study can contribute to technology management in broad energy research. From a technology management perspective, the study enriches the growing body of literature in using agent-based simulation methods to study technology adoption and post-adoption user behaviour. Using agent-based simulation as analytic models to study technology management has been booming since the late 1990s and one critique of the method is that it does not deal with real data and is therefore only for "toy problems" (Rand and Rust, 2011). Our research shows that an agent-based model can be partly based on real data (empirical survey data) and respond to real world problems. Our research outputs can provide the city council with energy policy/intervention implications on better managing the benefits of smart meters in smart meter deployment planning (e.g. facilitating energy consumers' learning of smart meters via informational interventions such as energy educational programmes, events or energy saving campaigns, as discussed in Section 5.2). The research has developed some useful residential energy consumer agent templates. These templates are reusable, and can be modified to study broad consumer issues in the energy market. The agent-based simulation model developed in the research form a basis of a smart meter deployment planning platform, which if further developed can become a useful energy decision tool for city councils in the UK to devise effective local energy policies/interventions. From the research we reported in this paper we conclude that, although it is not possible to perfectly replicate the real energy market, agent-based simulation as a novel approach is a very useful policy support tool for studying technology management and consumer issues in the energy market.

This model has a promising potential for further research. First, with the model we can carry out further experiments to investigate the effects of user learning on authoritative technology adoption. For example, we can investigate whether the levels of technology sophistication can influence user learning; whether different types of social networks can facilitate user learning, etc. Second, by addressing the limitations, we can further develop the model and extend the study. For example, we can enable energy consumers to change the archetypes based on more empirical observations, which can further improve the fidelity of these agents. Additionally, if we incorporate more GIS data into the model, we can investigate whether the authoritative installations of smart meters in council-owned properties can influence the voluntary adoption of smart meters in private properties.

**Acknowledgement**

The research was sponsored by the UK Engineering and Physical Sciences Research Council (Grant Ref: EP/G05956X/1). The authors would like to extend their thanks to their project collaborators in the University of Leeds

**Authors' Short-Bios:**

**Dr. Tao Zhang** joined the Department of Marketing as a Lecturer in Marketing and Sustainability August 2012. Prior to that he was a research fellow in the Intelligent Modelling & Analysis Research Group, School of Computer Science, University of Nottingham. He gained his PhD in Energy Economics from the Energy Policy Research Group, Judge Business School, University of Cambridge. His research interests are in the areas of energy economics and policy, energy consumer behaviour and innovation management, and agent-based modelling for the energy market. During his post-doc period in Nottingham, Dr. Zhang led the agent-based simulation part in an EPSRC funded research project *Future Energy Decision Making for Cities—Can Complexity Science Rise to the Challenge?* His publications appear in leading journals such as *Energy and Buildings*, *Energy Policy*, and the *Journal of Product Innovation Management*. He is a founding member of the Environmental and Energy Economics and Management research cluster in Birmingham Business School, University of Birmingham, an external advisor to the State Grid Corporation China, and an associate researcher of the ESRC Energy Policy Research Group.

**Dr. Peer-Olaf Siebers** is a Lecturer at the School of Computer Science, University of Nottingham, UK. He received his PhD from Cranfield University, UK in 2004. The central theme in his work is the development of human behaviour models which can be used to better represent people and their behaviours in Operational Research simulation. Dr. Siebers is a committed advocate of agent-based simulation. For his PhD he studied the impact of human performance variation on the accuracy of manufacturing system simulation models of manual assembly lines. His recent activities include the simulation of human resource management practices in retail, simulation of cargo screening processes at seaports, and monitoring the impact of different governmental and energy supplier interventions on consumer behaviour. He has published and presented in journals and conferences around the world.

**Prof. Uwe Aickelin** is a Professor of Computer Science at the University of Nottingham, where he is also the Head of School of Computer Science and a member of one of its main research groups: Intelligent Modelling & Analysis (IMA). Prof. Aickelin's previous roles include being an EPSRC Advanced Research Fellow, a member of EPSRC's Complexity SAT and ICT SAT and having acted as an EPSRC IDEAS Factory mentor and as a keynote speaker for various national and international conferences, including ICARIS, TradeTech and EPSRC's Early Career / leadership workshops for ICT. Prof. Aickelin's editorial duties include being an Associate Editor of the IEEE Transactions on Evolutionary Computation and an Editorial Board member of Evolutionary Intelligence. Prof. Aickelin has authored over 100 papers in international journals and conferences and participated in over 100 international Conference Programme Committees. Internationally, Prof. Aickelin works closely with Fudan University and Shenzhen University, where he has been appointed as a Special Professor.